\documentclass[sigconf]{acmart}
\usepackage{amsfonts}
\usepackage{amsmath}
\usepackage{booktabs}
\usepackage{float}
\usepackage{graphicx}
\usepackage{multirow}
\usepackage{url}
\usepackage{xcolor}
\usepackage{xspace}
\usepackage{caption}
\usepackage{subcaption}
\usepackage{kky}

\usepackage{xcolor}
\newcommand{\peter}[1]{\textbf{\textcolor{cyan}{peter: #1}}}

\newcommand{\attentionxml}{{\sf AttentionXML}\xspace}
\newcommand{\annexml}{{\sf AnnexML}\xspace}
\newcommand{\bonsai}{{\sf Bonsai}\xspace}
\newcommand{\discmec}{{\sf DiSMEC}\xspace}

\newcommand{\mlcseq}{{\sf MLC2seq}\xspace}
\newcommand{\parabel}{{\sf Parabel}\xspace}
\newcommand{\pfastrexml}{{\sf PfastreXML }\xspace}

\newcommand{\ppdsparse}{{\sf PPD-Sparse}\xspace}
\newcommand{\proxml}{{\sf ProXML}\xspace}
\newcommand{\sleec}{{\sf SLEEC}\xspace}
\newcommand{\slice}{{\sf SLICE}\xspace}
\newcommand{\xmlcnn}{{\sf XML-CNN}\xspace}
\newcommand{\xt}{{\sf eXtremeText}\xspace}

\newcommand{\tfidf}{{\sf PIFA}\xspace}

\newcommand{\slimrank}{{\sf X-Transformer}\xspace}

\newcommand{\best}[1]{\bf #1}
\newcommand{\eurlex}{{\sf Eurlex-4K}\xspace}
\newcommand{\wikis}{{\sf Wiki10-31K}\xspace}
\newcommand{\amzcat}{{\sf AmazonCat-13K}\xspace}

\newcommand{\wikil}{{\sf Wiki-500K}\xspace}
\newcommand{\ptoq}{{\sf Prod2Query-1M}\xspace}

\AtBeginDocument{%
  \providecommand\BibTeX{{%
    \normalfont B\kern-0.5em{\scshape i\kern-0.25em b}\kern-0.8em\TeX}}}

\copyrightyear{2020}
\acmYear{2020}
\setcopyright{rightsretained}
\acmConference[KDD '20]{Proceedings of the 26th ACM SIGKDD Conference on Knowledge Discovery and Data Mining}{August 23--27, 2020}{Virtual Event, CA, USA}
\acmBooktitle{Proceedings of the 26th ACM SIGKDD Conference on Knowledge Discovery and Data Mining (KDD '20), August 23--27, 2020, Virtual Event, CA, USA}
\acmDOI{10.1145/3394486.3403368}
\acmISBN{978-1-4503-7998-4/20/08}
\begin{document}
\fancyhead{}

%\title{Taminig Large-scale Deep Pretrained Transformers for \\ eXtreme Multi-label Text Classification}
\title{Taming Pretrained Transformers \\ for Extreme Multi-label Text Classification}

% \author{Anonymous}
% \email{Anonymous@corporation.com}
% \affiliation{Anonymous}

\author{Wei-Cheng Chang}
\affiliation{%
  \institution{Carnegie Mellon University}
  }

\author{Hsiang-Fu Yu}
\affiliation{%
  \institution{Amazon}
  }

\author{Kai Zhong}
\affiliation{%
  \institution{Amazon}
  }

\author{Yiming Yang}
\affiliation{%
  \institution{Carnegie Mellon University}
  }

\author{Inderjit S. Dhillon}
\affiliation{%
  \institution{Amazon \& UT Austin}
  }

\renewcommand{\shortauthors}{Chang, et al.}

\begin{abstract}
% define XMTC problem
We consider the extreme multi-label text classification~(XMC) problem: given an input text, return the most relevant labels  from a large label collection. For example, the input text could be a product description on Amazon.com and the labels could be product categories.
% Link XMTC as a "NLP" problem.
XMC is an important yet challenging problem in the NLP community.  
Recently, deep pretrained transformer models have achieved state-of-the-art performance on many NLP tasks including sentence classification, albeit with small label sets.
However, naively applying deep transformer models to the XMC problem leads to sub-optimal performance due to the large output space and the label sparsity issue.
In this paper, we propose \slimrank, the first scalable approach to fine-tuning deep transformer models for the XMC problem.
% should we dive in details? prob not?
The proposed method achieves new state-of-the-art results on four XMC benchmark datasets.
In particular, on a Wiki dataset with around 0.5 million labels, the prec@1 of \slimrank is $77.28\%$, a substantial improvement over state-of-the-art XMC approaches \parabel (linear) and \attentionxml (neural), which achieve $68.70\%$ and $76.95\%$ precision@1, respectively.
%\peter{todo: update because experiment results are not finalized yet}.
We further apply \slimrank to a product2query dataset from Amazon and gained 10.7\% relative improvement on prec@1 over \parabel. 
\end{abstract}




% %% A "teaser" image appears between the author and affiliation
% %% information and the body of the document, and typically spans the

\maketitle

\section{Introduction}

% define XMC problem
We are interested in the Extreme multi-label text classification (XMC) problem: given an input text instance, return the most relevant labels from an enormous label collection, where the number of labels could be in the millions or more.
% General form of XMC problem 
One can view the XMC problem as learning a score function $f: \Xcal \times \Ycal \rightarrow \RR$, that maps an (instance, label) pair $(\xb, \yb)$ to a score $f(\xb, \yb)$.
The function $f$ should be optimized such that highly relevant $(\xb, \yb)$ pairs have high scores, whereas the irrelevant pairs have low scores. 
% Many applications
Many real-world applications are in this form. 
For example, in E-commerce dynamic search advertising, $\xb$ represents an item and $\yb$ represents a bid query on the market~\cite{prabhu2014fastxml,prabhu2018parabel}.
In open-domain question answering, $\xb$ represents a question and $\yb$ represents an evidence passage containing the answer~\cite{lee2019latent,chang2020pretraining}.
In the PASCAL Large-Scale Hierarchical Text Classification (LSHTC) challenge, $\xb$ represents an article and $\yb$ represents a category of the Wikipedia hierarchical taxonomy~\cite{partalas2015lshtc}.

% XMC is a NLP problem
XMC is essentially a text classification problem on an industrial scale, which is one of the most important and fundamental topics in machine learning and natural language processing (NLP) communities.
% recent pretrained transformer
Recently, deep pretrained Transformers, e.g., BERT~\cite{devlin2018bert}, along with its many successors such as XLNet~\cite{yang2019xlnet} and RoBERTa~\cite{liu2019roberta}, have led to state-of-the-art performance on many tasks, such as question answering, part-of-speech tagging, information retrieval, and sentence classification with very few labels.
Deep pretrained Transformer models induce powerful token-level and sentence-level embeddings that can be rapidly fine-tuned on many downstream NLP problems by adding a task-specific lightweight linear layer on top of the transformer models.

% why cant directly finetune 24 layer BERT on XMC problem?
% two issues: gpu memory constraint and label sparsity
However, how to successfully apply Transformer models to XMC problems remains an open challenge, primarily due to the extremely large output space and severe label sparsity issues.
%it is incredibly challenging, to naively apply deep Transformer models on the XMC problem because of the large output space and label sparsity issues.
As a concrete example, Table~\ref{tab:model_memory} compares the model size (in terms of the number of model parameters) and GPU memory usage, when applying a 24-layer XLNet model to a binary classification problem (e.g., the MNLI dataset of
GLUE~\cite{wang2018glue}) versus its application to an XMC problem with 1 million labels. %Table~\ref{tab:model_memory} gives a detailed comparison. 
Note that the classifier for the MNLI problem and XMC problem has a model size of 2K and 1025M, respectively. 
This means that the latter is a much harder problem than the former from the model optimization point of view.
Additionally, in attempting to solve the XMC problem, we run out of GPU memory even for a single example mini-batch update.  
Table~\ref{tab:model_memory} gives the details of the GPU memory usage in the training stages of one forward pass, one backward pass and one optimization step, respectively.

\begin{table*}[!ht]
    \centering
\begin{tabular}{cccc|cccc}
    \toprule
    & \multicolumn{3}{c}{ XLNet-large model (\# params) } &  \multicolumn{4}{c}{(batch size, sequence length)=(1,128)} \\
    problem     & encoder   & classifier    & total     & load model & +forward & +backward & +optimizer step \\
    \midrule
    GLUE (MNLI) & 361 M     & 2 K           & 361 M      & 2169 MB & 2609 MB & 3809 MB & 6571 MB     \\
    XMC (1M)    & 361 M    & 1,025 M        & 1,386 M    & 6077 MB & 6537 MB & OOM & OOM \\
    \bottomrule
\end{tabular}
    \caption{On the left of are the model sizes (numbers of parameters) when applying the XLNet-large model to the MNLI problem vs. the XMC (1M) problem; on the right is the GPU memory usage (in megabytes) in solving the two problems, respectively. The results were obtained on a recent Nvidia 2080Ti GPU with 12GB memory. OOM stands for out-of-memory.}
    \label{tab:model_memory}
    \vspace{-1.0em}
\end{table*}

\begin{figure}[!ht]
    \centering
    \vspace{-.5em}
    \includegraphics[width=1.0\columnwidth]{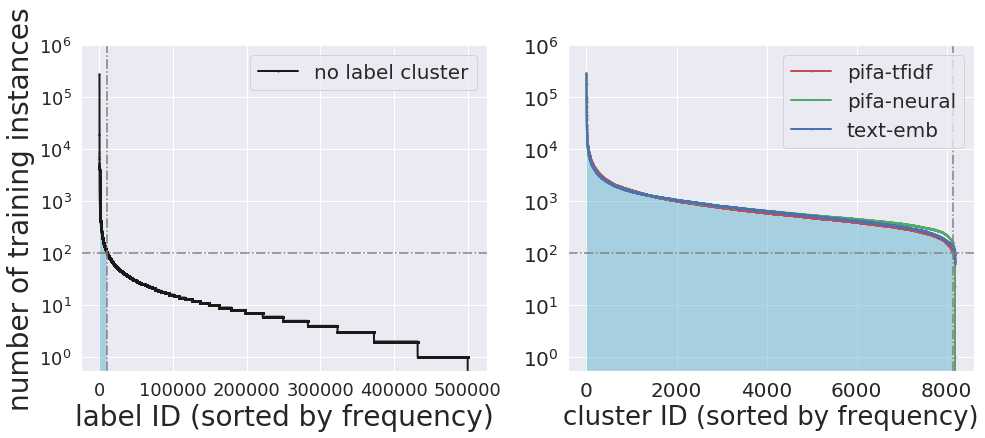}
    \vspace{-1.5em}
    \caption{On the left, \wikil shows a long-tail distribution of labels.
    Only $2.1\%$ of the labels have more than 100 training instances, as indicated by the cyan blue regime.
    On the right is the clusters distribution after our semantic label indexing based on different label representations;
    $99.4\%$ of the clusters have more than 100 training instances, which mitigates the data sparsity issue for fine-tuning of Transformer models.
    }
    \vspace{-0.25em}
    \label{fig:label-distr-wiki}
\end{figure}

% label sparsity
In addition to the computational challenges, the large output space in XMC is exacerbated by a severe label sparsity issue. The left part of 
Figure~\ref{fig:label-distr-wiki} illustrates the ``long-tailed'' label distribution in the \wikil data set~\cite{xmc_repo}.
Only $2\%$ of the labels have more than 100 training instances, while the remaining $98\%$ are long-tail labels with much fewer training instances. How to successfully fine-tune Transformer models with such sparsely labeled data is a tough question that has not been well-studied so far, to the best of our knowledge.
%This label sparsity issue further increases the difficulty of fine-tuning Transformer models for XMC, resulting in sub-optimal performance.

% ELMo style fine-tuning, more economic, but trading-off model power.  
Instead of fine-tuning deep Transformer models and dealing with the bottleneck classifier layer, an alternative is to use a more economical transfer learning paradigm as studied in the context of word2vec~\cite{mikolov2013distributed}, ELMo~\cite{peters2018deep}, and GPT~\cite{radford2018improving}.
For instance, ELMo uses a (bi-directional LSTM) model pretrained on large unlabeled text data to obtain contexualized word embeddings.
When applying ELMo on a downstream task, these word embeddings can be used as input without adaptation.
This is equivalent to freezing the ELMo encoder, and fine-tuning the downstream task-specific model on top of ELMo, which is much more efficient in terms of memory as well as computation.
However, such a benefit comes at the price of limiting the model capacity from adapting the encoder, as we will see in the experimental results in Section~\ref{sec:exp}.

% our contribution
In this paper, we propose \slimrank, %the first scalable approach that
a new approach that overcomes the aforementioned issues, with successful fine-tuning of deep Transformer models for the XMC problem.
% Transformer-XMC three component
%In a nutshell, 
\slimrank consists of a Semantic Label Indexing component, a Deep Neural Matching component, and an Ensemble Ranking component.
% Indexing
%Specifically, 
First, Semantic label Indexing~(SLI) decomposes the original intractable XMC problem into a set of feasible sub-problems of much smaller output space via label clustering, which mitigates the label sparsity issue as shown in the right part of Figure~\ref{fig:label-distr-wiki}.
% Matching
Second, the Deep Neural Matching component fine-tunes a Transformer model for each of the SLI-induced XMC sub-problems, resulting in a better mapping from the input text to the set of label clusters.
% Ranking
Finally, the Ensemble Ranking component is trained conditionally on the instance-to-cluster assignment and neural embedding from the Transformer, and is used to 
assemble scores derived from various SLI-induced sub-problems for further performance improvement.
  
In our experiments, the proposed \slimrank achieves new state-of-the-art results on four XMC benchmarks and leads to improvement on two real-would XMC applications.
% Wiki-500K
%Concretely, \slimrank achieves new state-of-the-art results on four XMC benchmark.
On a Wiki dataset with a half million labels, the precision@1 of \slimrank reaches $77.28\%$, a substantial improvement over the well-established hierarchical label tree approach \parabel~\cite{prabhu2018parabel} (i.e., $68.70\%$) and the competing deep learning method \attentionxml~\cite{you2019attentionxml} (i.e., $76.95\%$).
%, which achieve 68.70\% and 76.95\% precision@1, respectively.
% AMZ app
Furthermore, \slimrank also demonstrates great impact on the scalability of deep Transformer models in real-world large applications.
In our application of \slimrank to Amazon Product2Query problem that can be formulated as XMC, \slimrank significantly outperforms \parabel too. % in online experiments by a 12.1\% relative lift in conversion shares. 
The dataset, experiment code, models are available: \url{https://github.com/OctoberChang/X-Transformer}.

\section{Related Work and Background}
\label{sec:framework}

\iffalse
\begin{figure*}[!ht]
    \includegraphics[width=.8\textwidth]{figures/arch.png}
    \vspace{-0.5em}
    \caption{A diagram of the three-stage procedure of \slimrank for extreme multi-label classification}
    \label{fig_arch}
\end{figure*}
\fi

\subsection{Extreme Multi-label Classification}
\noindent \paragraph{\bf{Sparse Linear Models}}
% OVA
To overcome computational issues, most existing XMC algorithms use sparse TF-IDF features~(or slight variants), and leverage different partitioning techniques on the label space to reduce complexity.
For example, sparse linear one-vs-all~(OVA) methods such as DiSMEC \cite{babbar2017dismec}, \proxml~\cite{babbar2019data} and PPDSparse \cite{yen2017ppdsparse} explore parallelism to speed up the algorithm and reduce the model size by truncating model weights to encourage sparsity.
OVA approaches are also widely used as building blocks for many other approaches, for example, in \parabel \cite{prabhu2018parabel} and \slice~\cite{jain2019slice}, linear OVA classifiers with smaller output domains are used.

% label partition-based
The efficiency and scalability of sparse linear models can be further improved by incorporating different partitioning techniques on the label spaces.
For instance, 
\parabel~\cite{prabhu2018parabel} partitions the labels through a balanced 2-means label tree using label features constructed from the instances.
Recently, several approaches are proposed to improve \parabel. \bonsai~\cite{khandagale2019bonsai} relaxes two main constraints in \parabel: 1) allowing multi-way instead of binary partitionings of the label set at each intermediate node, and 2) removing strict balancing constraints on the partitions.
\slice~\cite{jain2019slice} considers building an approximate nearest neighbor (ANN) graph as an indexing structure over the labels. For a given instance, the relevant labels can be found quickly from the nearest neighbors of the instance via the ANN graph.

\noindent \paragraph{\bf{Deep Learning Approaches}}
Instead of using handcrafted TF-IDF features which are hard to optimize for different downstream XMC problems, deep learning approaches employ various neural network architectures to extract semantic embeddings of the input text.
\xmlcnn~\cite{liu2017deep} employs one-dimensional Convolutional neural networks along both sequence length and word embedding dimension for representing text input.
As a follow-up, \slice considers dense embedding from the supervised pre-trained \xmlcnn models as the input to its hierarchical linear models.
More recently, \attentionxml~\cite{you2019attentionxml} uses BiLSTMs and label-aware attention as the scoring function, and performs warm-up training of the models with hierarchical label trees.
In addition, \attentionxml consider various negative sampling strategies on the label space to avoid back-propagating the entire bottleneck classifier layer.

\subsection{Transfer Learning Approaches in NLP}
%\peter{I think the current logic flow seems better..}
Recently, the NLP community has witnessed a dramatic paradigm shift towards the ``pre-training then fine-tuning'' framework.
% BERT
One of the pioneering works is BERT~\cite{devlin2018bert}, whose pre-training objectives are masked token prediction and next sentence prediction tasks.
After pre-training on large-scale unsupervised corpora such as Wikipedia and BookCorpus, the Transformer model demonstrates vast improvement over existing state-of-the-art when fine-tuned on many NLP tasks such as the GLUE benchmark~\cite{wang2018glue}, named entity recognition, and question answering.
% XLNet and RoBERTa
More advanced variants of the pre-trained Transformer models include XLNet~\cite{yang2019xlnet} and RoBERTa~\cite{liu2019roberta}.
XLNet considers permutation language modeling as the pre-training objective and two-stream self-attention for target-aware token prediction. 
It is worth noting that the contextualized token embeddings extracted from XLNet also demonstrate competitive performance when fed into a task-specific downstream model on large-scale retrieval problems.
RoBERTa improves upon BERT by using more robust optimization with large-batch size update, and pre-training the model for longer till it truly converges.

However, transferring the success of these pre-trained Transformer models on the GLUE text classification to the XMC problem is non-trivial, as we illustrated in Table~\ref{tab:model_memory}.
Before the emergence of BERT-type end-to-end fine-tuning, the canonical way of transfer learning in NLP perhaps comes from the well-known Word2Vec~\cite{mikolov2013distributed} or GloVe~\cite{pennington2014glove} papers.
Word2vec is a shallow two-layer neural network that is trained to reconstruct the linguistic context of words.
GLoVe considers a matrix factorization objective to reconstruct the global word-to-word co-occurrence in the corpus.
A critical downside of Word2vec and GloVe is that the pre-trained word embeddings are not contextualized depending on the local surrounding word.
ELMo~\cite{peters2018deep} and GPT2~\cite{radford2018improving} instead present contextualized word embeddings by using large BiLSTM or Transformer models. 
After the models are pre-trained, transfer learning can be easily carried out by feeding these extracted word embeddings as input to the downstream task-specific models.
This is more efficient compared to the BERT-like end-to-end additional fine-tuning of the encoder, but comes at the expense of losing model expressiveness.
In the experimental results section, we show that using fixed word embeddings from universal pre-trained models such as BERT is not powerful enough for XMC problems.

\subsection{Amazon Applications}
\begin{figure}[!ht]
    \centering
        \includegraphics[width=0.9\linewidth]{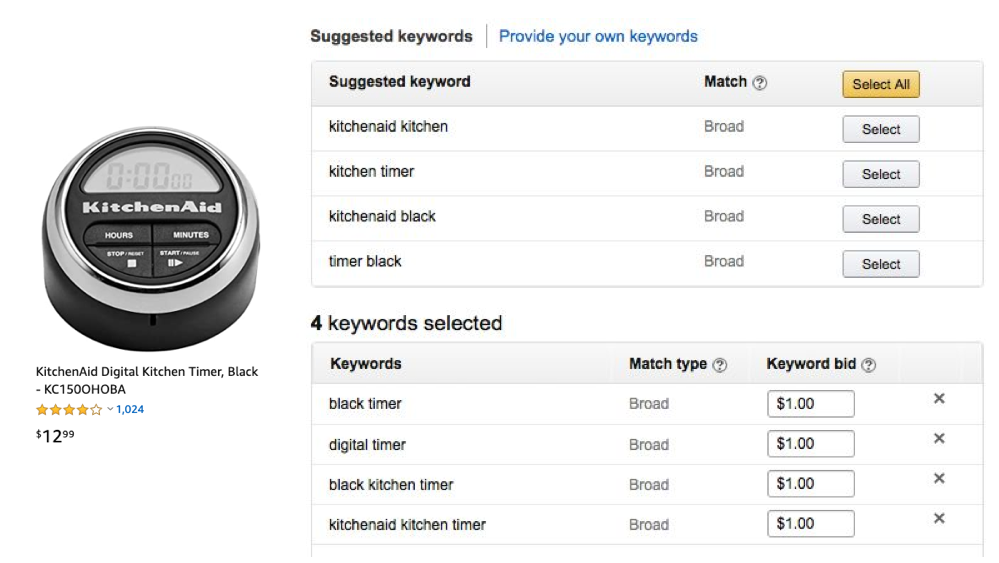}
        \caption{keyword recommendation system}
        \label{fig:real_apps}
\end{figure}

Many challenging problems at Amazon amount to finding relevant results from an enormous output space of potential candidates: for example, suggesting keywords to advertisers starting new campaigns on Amazon, predicting next queries a customer will type based on the previous queries he/she typed. Here we discuss keyword recommendation system for Amazon Sponsored Products, as illustrations in Fig.\ref{fig:real_apps}, and how it can be formulated as XMC problems. 

\noindent \paragraph{\bf Keyword recommendation system}
Keyword Recommendation Systems provide keyword suggestions for advertisers to create campaigns. In order to maximize the return of investment for the advertisers, the suggested keywords should be highly relevant to their products so that the suggestions can lead to conversion. An XMC model, when trained on an product-to-query dataset such as product-query customer purchase records, can suggest queries that are relevant to any given product by utilizing product information, like title, description, brand, etc. 

%\paragraph{\bf Search Indexing System}
%Indexing more and better term in Amazon's search engine leads to matching more products with greater relevance. An XMC model, when trained on an product-to-query dataset, generates relevant queries for a given product. Among these predicted queries, we find terms(words) that do not exist in the current search index but are relevant to the products.

\section{Proposed method: \slimrank}
\label{sec:main}

\begin{figure*}
    \centering
    \includegraphics[width=0.85\textwidth]{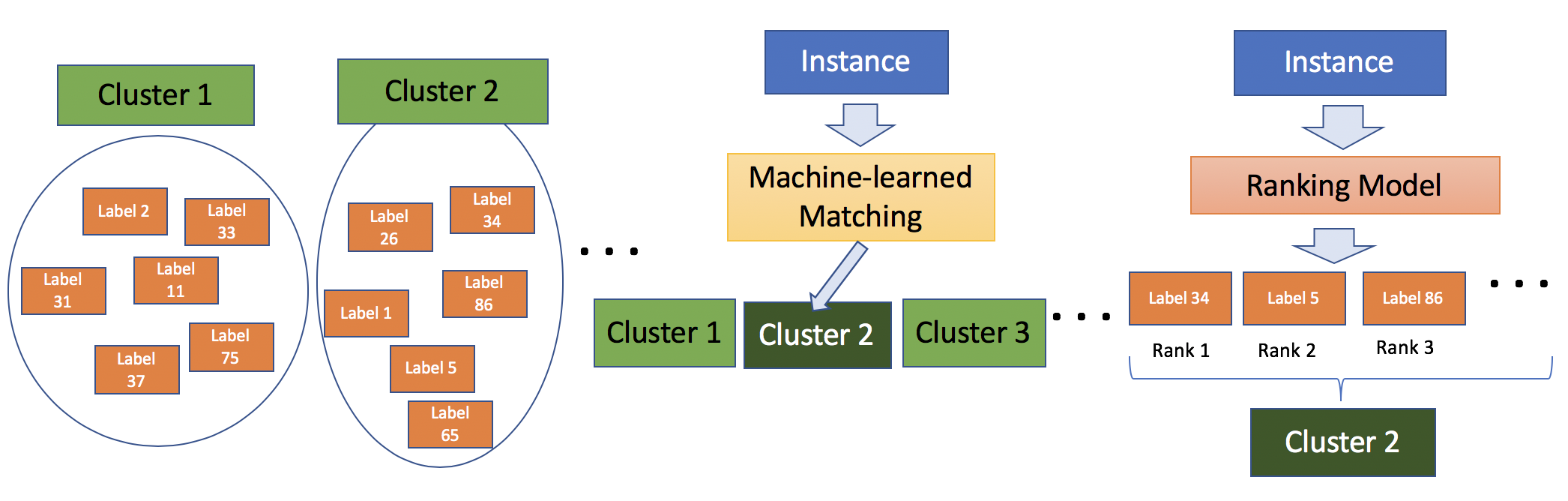}
    \vspace{-.75em}
    \caption{The proposed \slimrank framework. First, Semantic Label Indexing reduces the large output space.
    Transformers are then fine-tuned on the XMC sub-problem that maps instances to label clusters.
    Finally, linear rankers are trained conditionally on the clusters and Transformer's output in order to re-rank the labels within the predicted clusters.}
    \label{fig:model_pipeline}
\end{figure*}

\subsection{Problem Formulation}
\paragraph{\bf{Motivations}}
% define of score f
Given a training set $\Dcal=\{(\xb_i, \yb_i) | \xb_i \in \Xcal, \yb_i \in \{0,1\}^L, i=1,\ldots,N \}$, extreme multi-label classification aims to learn a scoring function $f$ that maps an input (or instance) $\xb_i$ and a label $l$ to a score $f(\xb_i, l) \in \RR$.
% train
The function $f$ should be optimized such that the score is high when $y_{il}=1$ (i.e., label $l$ is relevant to instance $\xb_i$) and the score is low when $y_{il}=0$.
A simple one-versus-all approach realizes the scoring function $f$ as
\begin{equation*}
    f(\xb,l) = \wb_l^T \phi(\xb)
\end{equation*}
where $\phi(\xb)$ represents an encoding and $\Wb=[\wb_1,\ldots, \wb_L]^T \in \RR^{L\times d}$ is the classifier bottleneck layer.
For convenience, we further define the top-$b$ prediction operator as
\begin{equation*}
    f_b(\xb) = \text{Top-b} \Big( \big[f(\xb,1),\ldots,f(\xb,L)\big] \Big) \in \{1,\ldots, L\},
\end{equation*}
where $f_b(\xb)$ is an index set containing the top-$b$ predicted labels.
As we pointed out in Table~\ref{tab:model_memory}, it is not only very difficult to fine-tune the Transformer encoders $\phi_{T}(\xb;\theta)$ together with the intractable classifier layer $\Wb$, but also extremely slow to compute the top-K predicted labels efficiently.

\paragraph{\bf{High-level Sketch}} 
To this end, we propose \slimrank as a practical solution to fine-tune deep Transformer models on XMC problems.
Figure~\ref{fig:model_pipeline} summarizes our proposed framework.

% clustering
In a nutshell, \slimrank decomposes the intractable XMC problem to a feasible sub-problem with a smaller output space, which is induced from semantic label indexing, which clusters the labels.
% matcher
We refer to this sub-problem as the neural matcher of the following form:
\begin{equation}
    g(\xb,k) = \wb_k^T \phi_{T}(\xb), \quad k=1,\ldots,K
\end{equation}
where $K$ is the number of clusters which is significantly smaller than the original intractable XMC problem of size $\Ocal(L)$.
% ranker
Finally, \slimrank currently uses a linear ranker that conditionally depends on the embedding of transformer models and its top-$b$ predicted clusters $g_b(\xb)$.
\begin{equation}
    f(\xb,l) =
    \begin{cases}
        \sigma\big( g(\xb, c_l), h(\xb, l) \big),   & \text{if $c_l \in g_b(\xb)$}, \label{eq:score_fn} \\
        -\infty,                                    & \text{otherwise}. 
    \end{cases}
\end{equation}
Here $c_l \in \{1,\ldots,K\}$ represents the cluster index of label $l$, $g(\xb, c_l)$ is the neural matcher realized by deep pre-trained Transformers, $h(\xb,l)$ is the linear ranker, and $\sigma()$ is a non-linear activation function to combine the final scores from $g$ and $h$. 
We now further introduce each of these three components in detail.

\subsection{Semantic Label Indexing}

Inducing latent clusters with semantic meaning brings several advantages to our framework.
We can perform a clustering of labels that can be represented by a label-to-cluster assignment matrix $\Cb \in \{0,1\}^{L \times K}$ where $c_{lk} = 1$ means label $l$ belongs to cluster $k$.
The number of clusters $K$ is typically set to be much smaller than the original label space $L$.
Deep Transformer models are fine-tuned on the induced XMC sub-problem where the output space is of size $K$, which significantly reduces the computational cost and avoids the label sparsity issue in Figure~\ref{fig:label-distr-wiki}.
Furthermore, the label clustering also plays a crucial role in the linear ranker $h(\xb,l)$.
For example, only labels within a cluster are used to construct negative instances for training the ranker.
In prediction, ranking is only performed for labels within a few clusters predicted by our deep Transformer models. 

Given a label representation, we cluster the $L$ labels hierarchically to form a hierarchical label tree with $K$ leaf nodes~\cite{prabhu2018parabel,jain2019slice,you2019attentionxml,khandagale2019bonsai}.
For simplicity, we consider binary balanced hierarchical trees~\cite{malinen2014balanced,prabhu2018parabel} as the default setting.
Due to the lack of a direct and informative representation of the labels, the indexing system for XMC may be noisy.
Fortunately, the instances in XMC are typically very informative. Therefore, we can utilize the rich information of the instances to build a strong matching system as well as a strong ranker to compensate for the indexing system.

\paragraph{\bf Label embedding via label text}
Given text information about labels, such as a short description of categories in the Wikipedia dataset or search queries on the Amazon shopping website, we can use this short text to represent the labels.
In this work, we use a pretrained XLNet~\cite{peters2018deep} to represent the words in the label.
The label embedding is the mean pooling of all XLNet word embeddings in the label text.
Specifically, the label embedding of label $l$ is 
\begin{equation*}
    \psi_{\text{text-emb}}(l)
    = \frac{1}{|\text{text}(l)|} \sum_{w \in text(l)} \phi_{\text{xlnet}}(w)
\end{equation*}
where $\phi_{xlnet}(w)$ is the hidden embedding of token $w$ in label $l$.

\paragraph{\bf Label embedding via embedding of positive instances}
The short text of labels may not contain sufficient information and is often ambiguous and noisy for some XMC datasets.
Therefore we can derive a label representation from embedding of its positive instances.
Specifically, the label embedding of label $l$ is
\begin{align*}
    \psi_{\text{pifa-tfidf}}(l) &= \vb_l / \|\vb_l\|,\; \vb_l = \sum_{i: y_{il} = 1} \phi_{\text{tf-idf}}(\xb_i), \quad l=1,\ldots,L, \\
    \psi_{\text{pifa-neural}}(l) &= \vb_l / \|\vb_l\|,\; \vb_l = \sum_{i: y_{il} = 1} \phi_{\text{xlnet}}(\xb_i), \quad l=1,\ldots,L.
\end{align*}
We refer to this type of label embedding as Positive Instance Feature Aggregation~(\tfidf), which is used in recent state-of-the-art XMC methods~\cite{prabhu2018parabel,jain2019slice,you2019attentionxml,khandagale2019bonsai}. 
Note that \slimrank is not limited by the above mentioned label representations; indeed
in applications where labels encode richer meta information such as a graph, we can use label representations derived from graph clustering and graph convolution.

\subsection{Deep Transformer as Neural Matcher}
After Semantic Label Indexing (SLI), the original intractable XMC problem morphs to a feasible XMC sub-problem with a much smaller output space of size~$K$.
See Table~\ref{tb:data} for the exact $K$ that we used for each XMC data set.
Specifically, the deep Transformer model now aims to map each text instance to the assigned relevant clusters.
The induced instance-to-cluster assignment matrix is
% \begin{equation*}
%     \Mb = \Yb \Cb = [\mb_1, \ldots, \mb_i, \ldots, \mb_N]^T \in \{0,1\}^{N \times K},
% \end{equation*}
$\Mb = \Yb \Cb = [\mb_1, \ldots, \mb_i, \ldots, \mb_N]^T \in \{0,1\}^{N \times K}$
where $\Yb \in \RR^{N \times L}$ is the original instance-to-label assignment matrix and $\Cb \in \RR^{L \times K}$ is the label-to-cluster assignment matrix provided by the SLI stage.
The goal now becomes fine-tuning deep Transformer models $g(\xb, k; \Wb,\theta)$ on $\{(\xb_i, \mb_i) | i=1,\ldots, N\}$ such that
\begin{align}
    \min_{\Wb,\theta} \quad& \frac{1}{N K} \sum_{i=1}^N \sum_{k=1}^K \max\big(0, 1 - \tilde{M}_{ik} g(\xb, k; \Wb,\theta) \big)^2, \label{eq:match-obj} \\
    \text{s.t.} \quad& g(\xb, k; \Wb,\theta) = \wb_k^T \phi_{\text{transformer}}(\xb), \nonumber
\end{align}
where $\tilde{\Mb}_{ik} = 2\Mb_{ik}-1 \in \{-1,1\}$, $\Wb=[\wb_1,\ldots,\wb_K]^T \in \RR^{K \times d}$, and $\phi_{\text{transformer}}(\xb) \in \RR^d$ is the embedding from the Transformers.  
We use the squared-hinge loss in the matching objective~\eqref{eq:match-obj} as it has shown better ranking performance as shown in~\cite{yen2017ppdsparse}.
Next, we discuss engineering optimizations and implementation details that considerably improve training efficiency and model performance.

\paragraph{\bf Pretrained Transformers}
We consider three state-of-the-art pre-trained Transformer-large-cased models (i.e., 24 layers with case-sensitive vocabulary) to fine-tune, namely BERT~\cite{devlin2018bert}, XLNet~\cite{yang2019xlnet}, and RoBERTa~\cite{liu2019roberta}.
The instance embedding $\phi(\xb)$ is the "[CLS]"-like hidden states from the last layer of BERT, RoBERTa and XLNet.
Computationally speaking, BERT and RoBERTa are similar while XLNet is nearly $1.8$ times slower.
In terms of performance on XMC tasks, we found RoBERTa and XLNet to be slightly better than BERT, but the gap is not as significant as in the GLUE benchmark. 
More concrete analysis is available in Section~\ref{sec:exp}. 

% float 16 vs float 32
It is possible to use Automatic Mixed Precision (AMP) between Float32 and Float16 for model fine-tuning, which can considerably reduce the model's GPU memory usage and training speed.
However, we used Float32 for all the experiments as our initial trials of training Transformers in AMP mode often led to unstable numerical results for the large-scale XMC dataset~\wikil.

\paragraph{\bf Input Sequence Length}
The time and space complexity of the Transformer scales quadratically with the input sequence length, i.e., $\Ocal(T^2)$~\cite{vaswani2017attention}, where $T=\text{len}(\xb)$ is the number of tokenized sub-words in the instance $\xb$.
Using smaller $T$ reduces not only the GPU memory usage that supports using larger batch size, but also increases the training speed.
For example, BERT first pre-trains on inputs of sequence length $128$ for $90\%$ of the optimization, and the remaining $10\%$ of optimization steps on inputs of sequence length $512$~\cite{devlin2018bert}. 
Interestingly, we observe that the model fine-tuned with sequence length $128$ v.s. sequence length $512$ does not differ significantly in the downstream XMC ranking performance.
Thus, we fix the input sequence length to be $T=128$ for model fine-tuning, which significantly speeds up the training time.
It would be interesting to see if we can bootstrap training the Transformer models from shorter sequence length and ramp up to larger sequence length (e.g., 32, 64, 128, 256), but we leave that as future work.

\paragraph{\bf Bootstrapping Label Clustering and Ranking}
After fine-tuning a deep Transformer model, we have powerful instance representation $\phi_{\text{transformer}}(\xb)$ that can be used to bootstrap semantic label clustering and ranking.
For label clustering, the embedding label $l$ can be constructed by aggregating the embeddings of its positive instances.
For ranking, the fine-tuned Transformer embedding can be concatenated with the sparse TF-IDF vector for better modeling power.
See details in the ablation study Table~\ref{tb:ablation}.

\begin{figure}[!tb]
    \centering
    \includegraphics[width=.98\columnwidth]{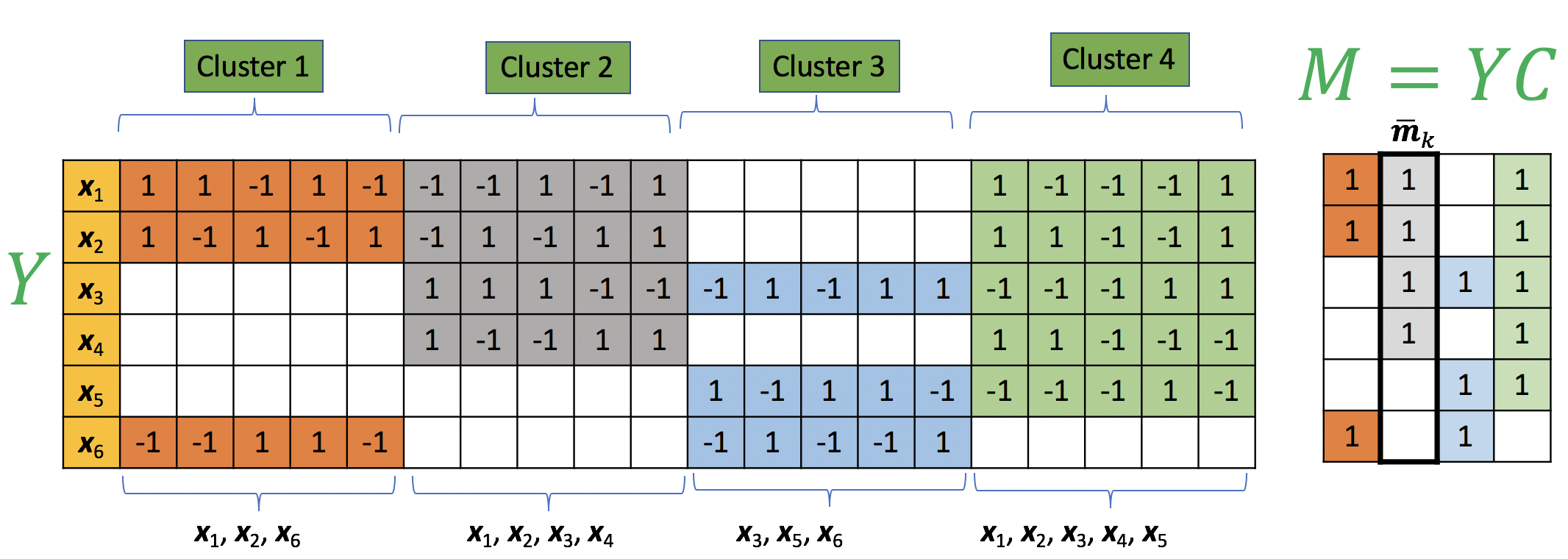}
    \vspace{-1em}
    \caption{\small Training rankers with the Teacher Forcing Negatives(TFN) strategy.
    For illustration, we have $N=6$ instances, $L=20$ labels, $K=4$ label clusters, and $\Mb \in \{0, 1\}^{6 \times 4}$ denotes the instance-to-cluster assignment matrix.
    For example, Cluster 1 with the orange color contains the first $5$ labels.
    The nonzeros of the first column of $\Mb$ correspond to $\{\xb_1,\xb_2,\xb_6\}$, which are instances with at least one positive label contained in Cluster 1.
    For each label in the first cluster, the ranker using Teacher Forcing Negatives~(TFN) only considers these three instances.
    Matcher-aware Negatives~(MAN) strategy is introduced in Section~\ref{sec:effrank} to further add improved hard negatives to enhance the TFN strategy.}
    \vspace{-.5em}
    \label{fig:ranker}
\end{figure}
\subsection{Ranking}\label{sec:effrank}
After the matching step, a small subset of label clusters is retrieved.
The goal of the ranker is to model the relevance between the instance and the labels from the retrieved clusters.
Formally, given a label $l$ and an instance $\xb$, we use a linear one-vs-all (OVA) classifier to parameterize the ranker $h(\xb,l)=\wb_l^T\phi(\xb)$ and train it with a binary loss.
For each label, naively estimating the weights $\wb_l$ based on all instances $\{(\xb_i,Y_{i,l})\}_{i=1}^N$ takes $\Ocal(N)$, which is too expensive.
Instead, we consider two sampling strategies that only include hard negative instances to reduce the computational complexity: Teacher Forcing Negatives~(TFN) and Matcher-aware Negatives~(MAN).

\textbf{Teacher Forcing Negatives~(TFN)}.
for each label $l$, we only include a subset of instances induced by the instance-to-cluster assignment matrix $\Mb = \Yb\Cb$.
In particular, in addition to the positive instances corresponding to the $l$-th label, we only include instances whose labels belong to the same cluster as the $l$-th label, i.e.,  $\{(\xb_i, y_{i,l}: i \in \{i:\Mb_{i,c_l}=1\} \}$. 
In Figure~\ref{fig:ranker}, we illustrate the TFN strategy with a toy example.
As the first five labels belong to Cluster $1$, and only $\{\xb_1,\xb_2,\xb_6\}$ contain a positive label within this cluster, we only consider this subset of instances to train a binary classifier for each of the first five labels. 

\textbf{Matcher-aware Negatives~(MAN)}.
The Teacher Forcing strategy only includes negative instances which are {\em hard} from the ``teacher'', i.e., the ground truth instance-to-clustering assignment matrix $\Mb$ used to train our neural matcher.
However, $\Mb$ is independent from the performance of our neural matcher.
Thus, training ranker with the TFN strategy alone might introduce an exposure bias issue, i.e., training-inference discrepancy. 
Instead, we also consider including matcher-aware hard negatives for each label. In particular, we can use the instance-to-cluster prediction matrix $\hat{\Mb} \in \{0, 1\}^{N\times K}$ from our neural matcher, where the nonzeros of the $i$-th row of $\hat{\Mb}$ correspond to the top-$b$ predicted clusters from $g_b(\xb_i)$.
In practice, we observe that a combination of TFN and MAN yields the best performance, i.e., using $\Mb' = \Yb\Cb + \hat{\Mb}$ to include hard negatives for each label. 
%instance we also consider augmenting the original $\Mb$ matrix with the label clusters predicted by the model $\hat{\Mb} \in \RR^{N \times K}$ such that $\Mb' = \Yb\Cb + \hat{\Mb}$, which further increases the number of ``hard matcher-aware'' negative examples. 
%We call this training strategy as Matcher-aware Negatives~(MAN).
See Table~\ref{tb:ablation} for a detailed Ablation study.

For the ranker input representation, we not only leverage the TF-IDF features $\phi_{\text{tf-idf}}(\xb)$, but also exploit the neural embeddings $\phi_{\text{neural}}(\xb)$ from either the pre-trained or fine-tuned Transformer model.
After the ranker is trained, the final ranking scores are computed via~\eqref{eq:score_fn}.
We can further ensemble the scores from different \slimrank models, which are trained on different semantic-aware label clusters or different pre-trained Transformer models such as BERT, RoBERTa and XLNet.

\section{Empirical Results}
\label{sec:exp}

The experiment code, including datasets and fine-tuned models are publicly available. \footnote{\url{https://github.com/OctoberChang/X-Transformer}}

\begin{table*}[!ht]
    \centering
    %\resizebox{1.0\textwidth}{!}{
    \begin{tabular}{c|rrrrrrrr}
    \toprule
        Dataset         & $n_{trn}$ & $n_{tst}$ & $|D_{\text{trn}}|$ & $|D_{\text{trn}}|$   & $L$       & $\bar{L}$ & $\bar{n}$ & $K$ \\
    \midrule
        \eurlex         & 15,449    & 3,865     & 19,166,707         & 4,741,799            & 3,956     & 5.30  & 20.79  & 64   \\
        \wikis          & 14,146    & 6,616     & 29,603,208         & 13,513,133           & 30,938    & 18.64 & 8.52   & 512  \\
        \amzcat         & 1,186,239 & 306,782   & 250,940,894        & 64,755,034           & 13,330    & 5.04  & 448.57 & 256  \\
        \wikil          & 1,779,881 & 769,421   & 1,463,197,965      & 632,463,513          & 501,070   & 4.75  & 16.86  & 8192 \\
    \bottomrule
    \end{tabular}
    %}
    \caption{Data Statistics.
    $n_{trn}, n_{tst}$ refer to the number of instances in the training and test sets, respectively.
    $|D_{\text{trn}}|, |D_{\text{tst}}|$ refer to the number of word tokens in the training and test corpus, respectively.
    $L$ is the number of labels,
    $\bar{L}$ the average number of labels per instance,
    $\bar{n}$ the average number of instances per label,
    and $K$ is the number of clusters.
    The four benchmark datasets are the same as \attentionxml~\cite{you2019attentionxml} for fair comparison.
    }
    \vspace{-1.25em}
    \label{tb:data}
\end{table*}

\subsection{Datasets and Preprocessing}
% xkc data
\paragraph{\bf XMC Benchmark Data}
We consider four multi-label text classification data sets used in \attentionxml~\cite{you2019attentionxml} for which we had access to the raw text representation, namely \eurlex, \wikis, \amzcat and \wikil.
Summary statistics of the data sets are given in Table \ref{tb:data}. We follow the training and test split of~\cite{you2019attentionxml} and set aside $10\%$ of the training instances as the validation set for hyperparameter tuning.

% a2q data
\paragraph{\bf Amazon Applications}

We consider an internal Amazon data set, %\footnote{ ``E-commerce'' is a placeholder to keep  anonymity during the review period. We will include the details later.},
namely \ptoq, which consists of 14 million instances (products) and 1 million labels~(queries) where the label is positive if a product is clicked at least once as a result of a search query.
We divide the data set into 12.5 million training samples, 0.8 million validation samples and 0.7 million testing samples. 

\subsection{Algorithms and Hyperparameters}

\paragraph{\bf Comparing Methods}
We compare our proposed \slimrank method to the most representative and state-of-the-art XMC methods including the embedding-based \annexml~\cite{tagami2017annexml}; one-versus-all \discmec~\cite{babbar2017dismec};  instance tree based \pfastrexml~\cite{jain2016extreme}; label tree based \parabel~\cite{prabhu2018parabel}, \xt~\cite{wydmuch2018no}, \bonsai~\cite{khandagale2019bonsai}; and deep learning based \xmlcnn~\cite{liu2017deep}, \attentionxml~\cite{you2019attentionxml} methods.
The results of all these baseline methods are obtained from~\cite[Table 3]{you2019attentionxml}.
For evaluation with other XMC approaches that have not released their code or are difficult to reproduce, we have a detailed comparison in Table~\ref{tab:more-exp}. 

\paragraph{\bf Evaluation Metrics}
We evaluate all methods with example-based ranking measures including Precision@k ($k=1,3,5$) and Recall@k ($k=1,3,5$), which are widely used in the XMC literature~\cite{prabhu2014fastxml,bhatia2015sparse,jain2016extreme,prabhu2018parabel,reddi2018stochastic}.

\paragraph{\bf Hyperparameters.}
For \slimrank, all hyperparameters are chosen from the held-out validation set.
The number of clusters are listed in Table~\ref{tb:data}, which are consistent with the \parabel setting for fair comparison. We consider the 24 layers cased models of BERT~\cite{devlin2018bert}, RoBERTa~\cite{liu2019roberta}, and XLNet~\cite{yang2019xlnet} using the Pytorch implementation from HuggingFace Transformers~\cite{Wolf2019HuggingFacesTS}\footnote{\url{https://github.com/huggingface/transformers}}.
For fine-tuning the Transformer models, we set the input sequence length to be $128$ for efficiency, and the batch size per GPU to be 16 along with gradient accumulation step of $4$, and use $4$ GPUs per model.
This together amounts to a batch size of $256$ in total.
We use Adam~\cite{kingma2014adam} with linear warmup scheduling as the optimizer where the learning rate is chosen from $\{4, 5, 6, 8\} \times 10^{-5}$.
Models are trained until convergence, which takes 1k, 1.4k, 20k, 50k optimization steps for \eurlex, \wikis, \amzcat, \wikil, respectively.
%\peter{no-op the best}.
%For \slimrank, we also consider warmup ratio of $0.2$.
%Generally speaking, we follow the default hyper-parameter as set in ~\cite{lin2017structured} and found it robust along with various semantic label indexing configurations and random seeds.

% main results
\begin{table*}[!ht]
    \centering
    %\resizebox{0.85\textwidth}{!}{
    \begin{tabular}{cccccccc}
        \toprule
        Methods & Prec@1 & Prec@3 & Prec@5 & Methods & Prec@1 & Prec@3 & Prec@5 \\
        \hline
        \hline
        \multicolumn{4}{c}{ Eurlex-4K }                             & \multicolumn{4}{c}{ Wiki10-31K } \\
        \midrule
        \annexml~\cite{tagami2017annexml}       & 79.66 & 64.94 & 53.52 & \annexml~\cite{tagami2017annexml}       & 86.46 & 74.28 & 64.20 \\
        \discmec~\cite{babbar2017dismec}        & 83.21 & 70.39 & 58.73 & \discmec~\cite{babbar2017dismec}        & 84.13 & 74.72 & 65.94 \\
        \pfastrexml~\cite{jain2016extreme}      & 73.14 & 60.16 & 50.54 & \pfastrexml~\cite{jain2016extreme}      & 83.57 & 68.61 & 59.10 \\
        \parabel~\cite{prabhu2018parabel}       & 82.12 & 68.91 & 57.89 & \parabel~\cite{prabhu2018parabel}       & 84.19 & 72.46 & 63.37 \\
        \xt~\cite{wydmuch2018no}                & 79.17 & 66.80 & 56.09 & \xt~\cite{wydmuch2018no}                & 83.66 & 73.28 & 64.51 \\
        \bonsai~\cite{khandagale2019bonsai}     & 82.30 & 69.55 & 58.35 & \bonsai~\cite{khandagale2019bonsai}     & 84.52 & 73.76 & 64.69 \\
        \mlcseq~\cite{nam2017maximizing}        & 62.77 & 59.06 & 51.32 &
        \mlcseq~\cite{nam2017maximizing}        & 80.79 & 58.59 & 54.66 \\
        \xmlcnn~\cite{liu2017deep}              & 75.32 & 60.14 & 49.21 & \xmlcnn~\cite{liu2017deep}              & 81.41 & 66.23 & 56.11 \\
        \attentionxml~\cite{you2019attentionxml}& 87.12 & 73.99 & 61.92 & \attentionxml~\cite{you2019attentionxml}& 87.47 & 78.48 & 69.37 \\
        \midrule
        $\phi_{\text{pre-xlnet}}$ + \parabel & 33.53 & 26.71 & 22.15 &
        $\phi_{\text{pre-xlnet}}$ + \parabel & 81.77 & 64.86 & 54.49 \\
        $\phi_{\text{tfidf}}$ + \parabel & 81.71 & 69.15 & 58.11 &
        $\phi_{\text{tfidf}}$ + \parabel & 84.27 & 73.20 & 63.66 \\
        $\phi_{\text{fnt-xlnet}} \oplus \phi_{\text{tfidf}}$ + \parabel & 84.09 & 71.50 & 60.12 &
        $\phi_{\text{fnt-xlnet}} \oplus \phi_{\text{tfidf}}$ + \parabel & 87.35 & 78.24 & 68.62 \\
        \slimrank                               & \best{87.22} & \best{75.12} & \best{62.90} &
        \slimrank                               & \best{88.51} & \best{78.71} & \best{69.62} \\
        \hline
        \hline
        \multicolumn{4}{c}{ AmazonCat-13K }     & \multicolumn{4}{c}{ Wiki-500K } \\
        \midrule
        \annexml~\cite{tagami2017annexml}       & 93.54 & 78.36 & 63.30 & \annexml~\cite{tagami2017annexml}       & 64.22 & 43.15 & 32.79 \\
        \discmec~\cite{babbar2017dismec}        & 93.81 & 79.08 & 64.06 & \discmec~\cite{babbar2017dismec}        & 70.21 & 50.57 & 39.68 \\
        \pfastrexml~\cite{jain2016extreme}      & 91.75 & 77.97 & 63.68 & \pfastrexml~\cite{jain2016extreme}      & 56.25 & 37.32 & 28.16 \\
        \parabel~\cite{prabhu2018parabel}       & 93.02 & 79.14 & 64.51 & \parabel~\cite{prabhu2018parabel}       & 68.70 & 49.57 & 38.64 \\
        \xt~\cite{wydmuch2018no}                & 92.50 & 78.12 & 63.51 & \xt~\cite{wydmuch2018no}                & 65.17 & 46.32 & 36.15 \\
        \bonsai~\cite{khandagale2019bonsai}     & 92.98 & 79.13 & 64.46 & \bonsai~\cite{khandagale2019bonsai}     & 69.26 & 49.80 & 38.83 \\
        \mlcseq~\cite{nam2017maximizing}        & 94.26 & 69.45 & 57.55 &
        \mlcseq~\cite{nam2017maximizing}        & - & - & - \\
        \xmlcnn~\cite{liu2017deep}              & 93.26 & 77.06 & 61.40 & \xmlcnn~\cite{liu2017deep}              & - & - & - \\
        \attentionxml~\cite{you2019attentionxml}& 95.92 & 82.41 & 67.31 & \attentionxml~\cite{you2019attentionxml}& 76.95 & \best{58.42} & \best{46.14} \\
        \midrule
        $\phi_{\text{pre-xlnet}}$ + \parabel & 80.96 & 63.92 & 50.72 &
        $\phi_{\text{pre-xlnet}}$ + \parabel & 31.83 & 20.24 & 15.76 \\
        $\phi_{\text{tfidf}}$ + \parabel & 92.81 & 78.99 & 64.31 &
        $\phi_{\text{tfidf}}$ + \parabel & 68.75 & 49.54 & 38.92 \\
        $\phi_{\text{fnt-xlnet}} \oplus \phi_{\text{tfidf}}$ + \parabel & 95.33 & 82.77 & 67.66 &
        $\phi_{\text{fnt-xlnet}} \oplus \phi_{\text{tfidf}}$ + \parabel & 75.57 & 55.12 & 43.31 \\
        \slimrank                               & \best{96.70} & \best{83.85} & \best{68.58} &
        \slimrank                               & \best{77.28} & 57.47 & 45.31 \\
        \bottomrule
    \end{tabular}
    %}
    \caption{Comparing \slimrank against state-of-the-art XMC methods on \eurlex, \wikis, \amzcat, and \wikil.
    The baselines' results are from~\cite[Table 3]{you2019attentionxml}.
    Note that \mlcseq and \xmlcnn are not scalable on \wikil.
    We also present linear baselines (\parabel) with three input representations.
    Specifically,
    $\phi_{\text{pre-xlnet}}$ denotes pre-trained XLNet embeddings,
    $\phi_{\text{tfidf}}$ denotes TF-IDF embeddings,
    $\phi_{\text{fnt-xlnet}} \oplus \phi_{\text{tfidf}}$ denotes fine-tuned XLNet embeddings concatenate with TF-IDF embeddings.
    }
    \vspace{-1.25em}
    \label{tb:main-results}
\end{table*}

\begin{table}[!ht]
    \centering
    %\begin{tabular}{c|ccccccc}
    \resizebox{1.0\columnwidth}{!}{
    \begin{tabular}{c|rrr|rrrrrrr}
        \toprule
        & \multicolumn{3}{c|}{Precision} &\multicolumn{3}{c}{Recall} \\
        Methods        & @1 & @5 & @10 & @1 & @5 & @10 \\
        %\hline  
        \midrule
        %$\phi_{\text{pre-bert}}(\xb)$ + \parabel & 7.54 & 4.95 & 3.76 & 2.02 & 5.38 & 7.45 \\
        %\hline
        %$\phi_{\text{tfidf}}(\xb)$ + \parabel & 50.72 & 28.94 & 19.98 & 24.74 & 51.02 & 60.40 \\ %\hline
        \slimrank & \textbf{10.7\%} & \textbf{7.4\%} & \textbf{6.6\%} & \textbf{12.0\%} & \textbf{4.9\%} & \textbf{2.8\%} \\
        \bottomrule
    \end{tabular}
    }
    \caption{Relative improvement over \parabel on the Prod2Query data set.}
    %\vspace{-.75em}
    \label{tb:prod2query_offline}
\end{table}

\subsection{Results on Public XMC Benchmark Data}

Table~\ref{tb:main-results} compares the proposed \slimrank with the most representative SOTA XMC methods on four benchmark datasets.
Following previous XMC works, we focus on top predictions by presenting Precision@k, where $k=1,3,5$.

The proposed \slimrank outperforms all XMC methods, except being slightly worse than \attentionxml in terms of P@3 and P@5 on the \wikil dataset.
We also compare \slimrank against linear baselines using \parabel model with three different input representations:
(1) $\phi_{\text{pre-xlnet}}$ denotes pretrained XLNet embeddings
(2) $\phi_{\text{tfidf}}$ denotes TF-IDF embeddings
(3) $\phi_{\text{fnt-xlnet}} \oplus \phi_{\text{tfidf}}$ denotes finetuned XLNet embeddings concatenated with TF-IDF embeeddings.
We clearly see that the performance of baseline~(1) is significantly worse. 
This suggests that the ELMo-style transfer learning, though efficient, is not powerful to achieve good performance for XMC problems.
%Both refers to the \elmo-style transfer learning that uses a fixed input embedding from pretrained networks as input to a SOTA XMC solver, where we consider \parabel.
The performance of baseline (2) is similar to that of \parabel, while baseline~(3) further improves performance due to the use of fine-tuned XLNet embeddings. 

\attentionxml~\cite{you2019attentionxml} is a very recent deep learning method that uses BiLSTM and label-aware attention layer to model the scoring function.
They also leverage hierarchical label trees to recursively warm-start the models and use hard negative sampling techniques to avoid using the entire classifier bottleneck layer.
Some of the techniques in \attentionxml are complementary to our proposed \slimrank, and it would be interesting to see how \slimrank can be improved from those techniques.

\subsection{\bf Results on Amazon Applications.}
%We applied the \slimrank model keyword recommendation system for Amazon ads campaign, where the advertisers bid keywords for their products.

%\noindent {\bf Offline Results.}
Recall that the Amazon data consists of 12 million products and 1 million queries along with product-query relevance. We treat queries as output labels and product title as input. We use the default \parabel method (using TFIDF features) as the baseline method and show \slimrank's relative improvement of precision and recall over the baseline in Table~\ref{tb:prod2query_offline}.

%\noindent {\bf Online Experimental Results.}
%We tested \slimrank using the Prod2Query data set in a keyword recommendation system~(KRS). The results show significant improvement over the current method: 6.6\% relative increase in  click share and 12.1\% relative increase in conversion share, where click share = (clicks from KRS-recommended keywords) / (clicks from all saved keywords) (and similarly for conversion share), where “all saved keywords” include keywords from KRS and keywords provided by advertisers themselves.

% ablation table
% \usepackage{multirow}
\begin{table*}[!ht]
    \centering
    \resizebox{1.0\textwidth}{!}{
    \begin{tabular}{cccccccccccc}
    \toprule
    \multirow{2}{*}{ Dataset } & \multirow{2}{*}{ Config. ID } & \multicolumn{4}{c}{ \slimrank Ablation Configuration } & \multicolumn{6}{c}{ Evaluation Metric } \\
    %\midrule
    & & indexing & matching & ranker input & negative-sampling  & P@1 & P@3 & P@5 & R@1 & R@3 & R@5 \\
    \midrule
    \multirow{10}{*}{ \eurlex }
    & 0 & pifa-tfidf    & BERT      & $\phi_{\text{tfidf}}(\xb)$                                  & TFN  & 83.93 & 70.59 & 58.69 & 17.05 & 42.08 & 57.14 \\
    & 1 & pifa-tfidf    & BERT      & $\phi_{\text{tfidf}}(\xb) \oplus \phi_{\text{neural}}(\xb)$ & TFN  & 85.02 & 71.83 & 59.87 & 17.21 & 42.79 & 58.30 \\
    & 2 & pifa-tfidf    & BERT      & $\phi_{\text{tfidf}}(\xb) \oplus \phi_{\text{neural}}(\xb)$ & TFN + MAN & 85.51 & 72.95 & 60.83 & 17.32 & 43.45 & 59.21 \\
    & 3 & pifa-tfidf    & RoBERTa   & $\phi_{\text{tfidf}}(\xb) \oplus \phi_{\text{neural}}(\xb)$ & TFN + MAN & 85.33 & 72.89 & 60.79 & 17.32 & 43.39 & 59.16 \\
    & 4 & pifa-tfidf    & XLNet     & $\phi_{\text{tfidf}}(\xb) \oplus \phi_{\text{neural}}(\xb)$ & TFN + MAN & 85.07 & 72.75 & 60.69 & 17.25 & 43.29 & 59.01 \\
    & 5 & pifa-neural   & XLNet     & $\phi_{\text{tfidf}}(\xb) \oplus \phi_{\text{neural}}(\xb)$ & TFN + MAN & 84.81 & 72.39 & 60.38 & 17.19 & 42.98 & 58.70 \\
    & 6 & text-emb      & XLNet     & $\phi_{\text{tfidf}}(\xb) \oplus \phi_{\text{neural}}(\xb)$ & TFN + MAN & 85.25 & 72.76 & 60.20 & 17.29 & 43.25 & 58.54 \\
    & 7 & all           & XLNet     & $\phi_{\text{tfidf}}(\xb) \oplus \phi_{\text{neural}}(\xb)$ & TFN + MAN & 86.55 & 74.24 & 61.96 & 17.54 & 44.16 & 60.24 \\
    & 8 & pifa-neural   & all       & $\phi_{\text{tfidf}}(\xb) \oplus \phi_{\text{neural}}(\xb)$ & TFN + MAN & 85.92 & 73.43 & 61.53 & 17.40 & 43.69 & 59.86 \\
    & 9 & all           & all       & $\phi_{\text{tfidf}}(\xb) \oplus \phi_{\text{neural}}(\xb)$ & TFN + MAN & \textbf{87.22} & \textbf{75.12} & \textbf{62.90} & \textbf{17.69} & \textbf{44.73} & \textbf{61.17} \\
    \midrule
    \multirow{10}{*}{ \wikil }
    & 0 & pifa-tfidf    & BERT      & $\phi_{\text{tfidf}}(\xb)$                                  & TFN & 69.52 & 49.87 & 38.71 & 22.30 & 40.62 & 48.65 \\
    & 1 & pifa-tfidf    & BERT      & $\phi_{\text{tfidf}}(\xb) \oplus \phi_{\text{neural}}(\xb)$ & TFN & 71.90 & 51.58 & 40.10 & 23.27 & 42.14 & 50.42 \\
    & 2 & pifa-tfidf    & BERT      & $\phi_{\text{tfidf}}(\xb) \oplus \phi_{\text{neural}}(\xb)$ & TFN + MAN & 74.68 & 53.64 & 41.50 & 24.56 & 44.26 & 52.50 \\
    & 3 & pifa-tfidf    & RoBERTa   & $\phi_{\text{tfidf}}(\xb) \oplus \phi_{\text{neural}}(\xb)$ & TFN + MAN & 75.40 & 54.32 & 42.06 & 24.85 & 44.93 & 53.30 \\
    & 4 & pifa-tfidf    & XLNet     & $\phi_{\text{tfidf}}(\xb) \oplus \phi_{\text{neural}}(\xb)$ & TFN + MAN & 75.45 & 54.50 & 42.24 & 24.81 & 45.00 & 53.44 \\
    & 5 & pifa-neural   & XLNet     & $\phi_{\text{tfidf}}(\xb) \oplus \phi_{\text{neural}}(\xb)$ & TFN + MAN & 76.34 & 55.50 & 43.04 & 25.15 & 45.88 & 54.53 \\
    & 6 & text-emb      & XLNet     & $\phi_{\text{tfidf}}(\xb) \oplus \phi_{\text{neural}}(\xb)$ & TFN + MAN & 74.12 & 52.85 & 40.53 & 24.18 & 43.30 & 50.98 \\
    & 7 & all           & XLNet     & $\phi_{\text{tfidf}}(\xb) \oplus \phi_{\text{neural}}(\xb)$ & TFN + MAN & 75.85 & 56.08 & 44.24 & 24.80 & 46.36 & 56.35 \\
    & 8 & pifa-neural   & all       & $\phi_{\text{tfidf}}(\xb) \oplus \phi_{\text{neural}}(\xb)$ & TFN + MAN & \textbf{77.44} & 56.84 & 44.37 & 25.61 & 47.18 & 56.55 \\
    & 9 & all           & all       & $\phi_{\text{tfidf}}(\xb) \oplus \phi_{\text{neural}}(\xb)$ & TFN + MAN & 77.28 & \textbf{57.47} & \textbf{45.31} & \textbf{25.48} & \textbf{47.82} & \textbf{57.95} \\
    \bottomrule
\end{tabular}
    }
    \caption{Ablation study of \slimrank on \eurlex and \wikil data sets.
    We outline four take away messages: (1) Config. ID$=\{0,1,2\}$ demonstrates better performance by using Matcher-aware Negatives (MAN) and Neural embedding for training the rankers; (2) Config. ID$=\{2,3,4\}$ suggests that, performance-wise, XLNet is similar to RoBERTa, and slightly better than BERT; (3) Config. ID=$\{4, 5,6\}$ manifests the importance of label clusters induced from different label representations. (4) Config. ID=$\{7,8,9\}$ indicates the effect of ensembling various configuration of the models.}
    \vspace{-1.25em}
    \label{tb:ablation}

\end{table*}

\subsection{\bf Ablation Study}
We carefully conduct an ablation study of \slimrank as shown in Table~\ref{tb:ablation}.
We analyze the \slimrank framework in terms of its four components: indexing, matching, ranker input representation, and training negative-sampling training algorithm.
The configuration Index $9$ represents the final best configuration as reported in Table~\ref{tb:main-results}.
There are four takeaway messages from this ablation study, and we describe them in the following four paragraphs.

\paragraph{\bf{Ranker Representation and Training}}
Config. ID $0,1,2$ shows the effect of input representation and training strategy for the ranking.
The benefit of using instance embedding from fine-tuned transformers can be seen from config. ID $0$ to $1$.
In addition, from ID $1$ to $2$, we observe that using Teacher Forcing Negatives (TFN) is not enough for training the ranker, as it could suffer from the exposure bias of only using the ground truth clustering assignment, but ignores the hard negatives mistakenly produced by the Transformer models.
Note that techniques such as adding Matcher-aware negatives (MAN) from previous model's prediction to bootstrap the next level's model training is also used in \attentionxml~\cite{you2019attentionxml}.

\paragraph{\bf{Different Transformer Models.}}
Next, we analyze how the three different Transformer models (i.e., BERT, RoBERTa, XLNet) affect the performance, as shown in Config. ID $2,3,4$.
For \wikil, we observe that the XLNet and RoBERTa are generally more powerful than the BERT models. 
On the other hand, such an advantage is not clear for \eurlex, possibly due to the nature of the data set. 

\paragraph{\bf{Label Representation for Clustering}}
The importance of different label representation for clustering is demonstrated in Config. ID $4,5,6$.
For \eurlex, we see that using label text embedding as representation (i.e. text-emb) leads to the strong performance compared to pifa-tfidf (id 4) and pifa-neural (id 5).
In contrast, pifa-tfidf becomes the best performing representation on the \wikil dataset.
This phenomenon could be due to the label text of \wikil being more noisy compared to \eurlex, which deteriorates the label clustering results on \wikil.

%TF-IDF essentially re-ranks the labels within the retrieved clusters by word matching with the queried documents without learning any model. We see a considerable performance drop from training linear models to the TF-IDF word matching in the re-ranking stage.
%This finding suggests the importance of learning more powerful parametric models in the ranking stage; in the future we plan to go beyond linear to neural models for ranking.

\paragraph{\bf{Ensemble Ranking}}
Finally, we show the advantage of ensembing prediction from different models as shown in Config. ID $7,8,9$.
For \eurlex, combining predictions from different label representations (ID 7) is better than from different Transformer models (ID 8).
Combining all (ID 9) leads to our final model, \slimrank.

\begin{table*}[!ht]
    \centering
    \resizebox{1.0\textwidth}{!}{
    \begin{tabular}{ccrrr||ccrrr}
        \toprule
        \multicolumn{5}{c}{\eurlex}  &    \multicolumn{5}{c}{\wikil} \\
        \midrule
        \multirow{3}{*}{Method} & \multirow{3}{*}{Source} & \multicolumn{3}{c||}{Relative Improvement} &
        \multirow{3}{*}{Method} & \multirow{3}{*}{Source} & \multicolumn{3}{c}{Relative Improvement}    \\
        & & \multicolumn{3}{c||}{over Parabel (\%)} & & & \multicolumn{3}{c}{over Parabel (\%)} \\
        & &   Prec@1 & Prec@3 & Prec@5 & & &   Prec@1 & Prec@3 & Prec@5 \\
        \hline
        \hline
        
        \slimrank   &    Table~\ref{tb:main-results}            &\best{+6.27\%}     &\best{+9.08\%}     &\best{+8.55\%} &
        \slimrank   &    Table~\ref{tb:main-results}            &\best{+12.49\%}    &\best{+15.94\%}    &\best{+17.26\%}      \\
        \slice      &    \cite[Table 2]{jain2019slice}          &      +4.27\%      &      +3.34\%      &      +3.11\% &
        \slice      &    \cite[Table 2]{jain2019slice}          &      +5.53\%      &      +7.02\%      &      +7.56\%       \\
        GLaS        &    \cite[Table 3]{guo2019breaking}        &      -5.18\%      &      -5.48\%      &      -5.34\% &
        GLaS        &    \cite[Table 3]{guo2019breaking}        &      +4.77\%      &      +3.37\%      &      +4.27\%      \\
        \proxml     &    \cite[Table 5]{babbar2019data}         &      +3.86\%      &      +2.90\%      &      +2.43\% &
        \proxml     &    \cite[Table 5]{babbar2019data}         &      +2.22\%      &      +0.82\%      &      + 2.92\%       \\
        \ppdsparse  &    \cite[Table 2]{prabhu2018parabel}      &      +1.92\%      &      +2.93\%      &      +2.92\% &
        \ppdsparse  &    \cite[Table 2]{prabhu2018parabel}      &      +2.39\%      &      +2.33\%      &      + 2.88\%       \\
        \sleec      &    \cite[Table 2]{khandagale2019bonsai}   &      -3.53\%      &      -6.40\%      &      -9.04\% &
        \sleec      &    \cite[Table 2]{khandagale2019bonsai}   &     -29.84\%      &      -40.73\%     &      -45.08\%       \\
    \bottomrule
    \end{tabular}
    }
    \caption{Comparison of Relative Improvement over \parabel. The relative improvement for each state-of-the-art~(SOTA) method is computed based on the metrics reported from its original paper as denoted in the Source column.
    }
    %The proposed \slimrank outperforms  all other SOTA approaches on two commonly used benchmark datasets: \eurlex and \wikil.}
    \label{tab:more-exp}
    %\vspace{-0.75em}
\end{table*}

\subsection{Cross-Paper Comparisons}
\label{sec:more-exp}

%Many XMC approaches have been recently proposed.
%While most of them contain an empirical comparison on a few commonly used datasets, such
%as \eurlex and \wikil, the evaluation metric of the same method on the ``same dataset'' varies from paper to paper.
Many XMC approaches have been proposed recently.
However, it is sometimes difficult to compare metrics directly from different papers.
For example, the P@1 of \parabel on \wikil is 59.34\% in~\cite[Table 2]{jain2019slice} and 68.52\% in~\cite[Table 2]{prabhu2018parabel}, but we see 68.70\% in Table~\ref{tb:main-results}.
The inconsistency may be due to differences in data processing, input representation, or other reasons.
We propose an approach to calibrate these numbers so that various methods can be compared in a more principled way.
In particular, for each metric $m(\cdot)$, we use the relative improvement over a common anchor method, which is set to be \parabel as it is widely used in the literature.
For a competing method X with a metric $m(\text{X})$ on a data set reported in a paper, we can compute the relative improvement over \parabel as follows: $\frac{m(\text{X}) - m(\text{\parabel})}{m(\text{\parabel})} \times 100 \%$,
where $m(\text{\parabel})$ is the metric obtained by \parabel on the same data set in the same paper.
Following the above approach, we include a variety of XMC approaches in our comparison.
We report the relative improvement of various methods on two commonly used data sets, \eurlex and \wikil, in Table~\ref{tab:more-exp}.
We can clearly observe that \slimrank brings the most significant improvement over \parabel and \slice.

\section{Conclusions}
\label{sec:conclusion}

In this paper, we propose \slimrank, the {\em first} scalable framework to fine-tune Deep Transformer models that improves state-of-the-art XMC methods on four XMC benchmark data sets.
We further applied \slimrank to a real-life application, product2query prediction, showing significant improvement over the competitive linear models, \parabel.%keyword recommendation system for ads at Amazon.
%Online experiments show significant improvement over the current system, e.g., \slimrank obtained 12.1\% relative lift in conversion share in the keyword recommendation system. 
%The novel semantic label indexing stage endows heterogeneous label partitions that bootstrap various BERT models, resulting in a powerful ensemble model for XMC problem.
%Quantitatively, on the \wikil dataset, the precision@1 is increased from $60.91\%$ to $67.87\%$ when comparing \slimrank to the strong XMC method \parabel. This amounts to a $11.43\%$ relative improvement over \parabel, which is indeed significant compared to the recent state-of-the-art approach \slice which has $5.53\%$ relative improvement over \parabel. 

% \begin{acks}
% \end{acks}

\bibliographystyle{ACM-Reference-Format}
\bibliography{sdp}


\begin{thebibliography}{32}

%%% the Web address.

\ifx \showCODEN    \undefined \def \showCODEN     #1{\unskip}     \fi
\ifx \showDOI      \undefined \def \showDOI       #1{#1}\fi
\ifx \showISBNx    \undefined \def \showISBNx     #1{\unskip}     \fi
\ifx \showISBNxiii \undefined \def \showISBNxiii  #1{\unskip}     \fi
\ifx \showISSN     \undefined \def \showISSN      #1{\unskip}     \fi
\ifx \showLCCN     \undefined \def \showLCCN      #1{\unskip}     \fi
\ifx \shownote     \undefined \def \shownote      #1{#1}          \fi
\ifx \showarticletitle \undefined \def \showarticletitle #1{#1}   \fi
\ifx \showURL      \undefined \def \showURL       {\relax}        \fi
% invisible to TeX
\providecommand\bibfield[2]{#2}
\providecommand\bibinfo[2]{#2}
\providecommand\natexlab[1]{#1}
\providecommand\showeprint[2][]{arXiv:#2}

\bibitem[\protect\citeauthoryear{Babbar and Sch{\"o}lkopf}{Babbar and
  Sch{\"o}lkopf}{2017}]%
        {babbar2017dismec}
\bibfield{author}{\bibinfo{person}{Rohit Babbar} {and}
  \bibinfo{person}{Bernhard Sch{\"o}lkopf}.} \bibinfo{year}{2017}\natexlab{}.
\newblock \showarticletitle{DiSMEC: distributed sparse machines for extreme
  multi-label classification}. In \bibinfo{booktitle}{\emph{WSDM}}.
\newblock


\bibitem[\protect\citeauthoryear{Babbar and Sch{\"o}lkopf}{Babbar and
  Sch{\"o}lkopf}{2019}]%
        {babbar2019data}
\bibfield{author}{\bibinfo{person}{Rohit Babbar} {and}
  \bibinfo{person}{Bernhard Sch{\"o}lkopf}.} \bibinfo{year}{2019}\natexlab{}.
\newblock \showarticletitle{Data scarcity, robustness and extreme multi-label
  classification}.
\newblock \bibinfo{journal}{\emph{Machine Learning}} (\bibinfo{year}{2019}),
  \bibinfo{pages}{1--23}.
\newblock


\bibitem[\protect\citeauthoryear{Bhatia, Jain, Kar, Varma, and Jain}{Bhatia
  et~al\mbox{.}}{2015}]%
        {bhatia2015sparse}
\bibfield{author}{\bibinfo{person}{Kush Bhatia}, \bibinfo{person}{Himanshu
  Jain}, \bibinfo{person}{Purushottam Kar}, \bibinfo{person}{Manik Varma},
  {and} \bibinfo{person}{Prateek Jain}.} \bibinfo{year}{2015}\natexlab{}.
\newblock \showarticletitle{Sparse local embeddings for extreme multi-label
  classification}. In \bibinfo{booktitle}{\emph{NIPS}}.
\newblock


\bibitem[\protect\citeauthoryear{Chang, Yu, Chang, Yang, and Kumar}{Chang
  et~al\mbox{.}}{2020}]%
        {chang2020pretraining}
\bibfield{author}{\bibinfo{person}{Wei-Cheng Chang}, \bibinfo{person}{Felix~X.
  Yu}, \bibinfo{person}{Yin-Wen Chang}, \bibinfo{person}{Yiming Yang}, {and}
  \bibinfo{person}{Sanjiv Kumar}.} \bibinfo{year}{2020}\natexlab{}.
\newblock \showarticletitle{Pre-training Tasks for Embedding-based Large-scale
  Retrieval}. In \bibinfo{booktitle}{\emph{International Conference on Learning
  Representations}}.
\newblock


\bibitem[\protect\citeauthoryear{Devlin, Chang, Lee, and Toutanova}{Devlin
  et~al\mbox{.}}{2019}]%
        {devlin2018bert}
\bibfield{author}{\bibinfo{person}{Jacob Devlin}, \bibinfo{person}{Ming-Wei
  Chang}, \bibinfo{person}{Kenton Lee}, {and} \bibinfo{person}{Kristina
  Toutanova}.} \bibinfo{year}{2019}\natexlab{}.
\newblock \showarticletitle{Bert: Pre-training of deep bidirectional
  transformers for language understanding}. In
  \bibinfo{booktitle}{\emph{Proceedings of the 2019 Conference of the North
  {A}merican Chapter of the Association for Computational Linguistics
  (NAACL)}}.
\newblock


\bibitem[\protect\citeauthoryear{Guo, Mousavi, Wu, Holtmann-Rice, Kale, Reddi,
  and Kumar}{Guo et~al\mbox{.}}{2019}]%
        {guo2019breaking}
\bibfield{author}{\bibinfo{person}{Chuan Guo}, \bibinfo{person}{Ali Mousavi},
  \bibinfo{person}{Xiang Wu}, \bibinfo{person}{Daniel~N Holtmann-Rice},
  \bibinfo{person}{Satyen Kale}, \bibinfo{person}{Sashank Reddi}, {and}
  \bibinfo{person}{Sanjiv Kumar}.} \bibinfo{year}{2019}\natexlab{}.
\newblock \showarticletitle{Breaking the Glass Ceiling for Embedding-Based
  Classifiers for Large Output Spaces}. In \bibinfo{booktitle}{\emph{Advances
  in Neural Information Processing Systems}}. \bibinfo{pages}{4944--4954}.
\newblock


\bibitem[\protect\citeauthoryear{Jain, Balasubramanian, Chunduri, and
  Varma}{Jain et~al\mbox{.}}{2019}]%
        {jain2019slice}
\bibfield{author}{\bibinfo{person}{Himanshu Jain}, \bibinfo{person}{Venkatesh
  Balasubramanian}, \bibinfo{person}{Bhanu Chunduri}, {and}
  \bibinfo{person}{Manik Varma}.} \bibinfo{year}{2019}\natexlab{}.
\newblock \showarticletitle{{Slice}: Scalable Linear Extreme Classifiers
  Trained on 100 Million Labels for Related Searches}. In
  \bibinfo{booktitle}{\emph{Proceedings of the Twelfth ACM International
  Conference on Web Search and Data Mining}}. ACM, \bibinfo{pages}{528--536}.
\newblock


\bibitem[\protect\citeauthoryear{Jain, Prabhu, and Varma}{Jain
  et~al\mbox{.}}{2016}]%
        {jain2016extreme}
\bibfield{author}{\bibinfo{person}{Himanshu Jain}, \bibinfo{person}{Yashoteja
  Prabhu}, {and} \bibinfo{person}{Manik Varma}.}
  \bibinfo{year}{2016}\natexlab{}.
\newblock \showarticletitle{Extreme multi-label loss functions for
  recommendation, tagging, ranking \& other missing label applications}. In
  \bibinfo{booktitle}{\emph{KDD}}.
\newblock


\bibitem[\protect\citeauthoryear{Khandagale, Xiao, and Babbar}{Khandagale
  et~al\mbox{.}}{2019}]%
        {khandagale2019bonsai}
\bibfield{author}{\bibinfo{person}{Sujay Khandagale}, \bibinfo{person}{Han
  Xiao}, {and} \bibinfo{person}{Rohit Babbar}.}
  \bibinfo{year}{2019}\natexlab{}.
\newblock \showarticletitle{Bonsai-Diverse and Shallow Trees for Extreme
  Multi-label Classification}.
\newblock \bibinfo{journal}{\emph{arXiv preprint arXiv:1904.08249}}
  (\bibinfo{year}{2019}).
\newblock


\bibitem[\protect\citeauthoryear{Kingma and Ba}{Kingma and Ba}{2014}]%
        {kingma2014adam}
\bibfield{author}{\bibinfo{person}{Diederik Kingma} {and}
  \bibinfo{person}{Jimmy Ba}.} \bibinfo{year}{2014}\natexlab{}.
\newblock \showarticletitle{Adam: A method for stochastic optimization}. In
  \bibinfo{booktitle}{\emph{Proceedings of the International Conference on
  Learning Representations}}.
\newblock


\bibitem[\protect\citeauthoryear{Lee, Chang, and Toutanova}{Lee
  et~al\mbox{.}}{2019}]%
        {lee2019latent}
\bibfield{author}{\bibinfo{person}{Kenton Lee}, \bibinfo{person}{Ming-Wei
  Chang}, {and} \bibinfo{person}{Kristina Toutanova}.}
  \bibinfo{year}{2019}\natexlab{}.
\newblock \showarticletitle{Latent retrieval for weakly supervised open domain
  question answering}. In \bibinfo{booktitle}{\emph{Proceedings of the 57th
  Annual Meeting of the Association for Computational Linguistics (ACL)}}.
\newblock


\bibitem[\protect\citeauthoryear{Liu, Chang, Wu, and Yang}{Liu
  et~al\mbox{.}}{2017}]%
        {liu2017deep}
\bibfield{author}{\bibinfo{person}{Jingzhou Liu}, \bibinfo{person}{Wei-Cheng
  Chang}, \bibinfo{person}{Yuexin Wu}, {and} \bibinfo{person}{Yiming Yang}.}
  \bibinfo{year}{2017}\natexlab{}.
\newblock \showarticletitle{Deep learning for extreme multi-label text
  classification}. In \bibinfo{booktitle}{\emph{Proceedings of the 40th
  International ACM SIGIR Conference on Research and Development in Information
  Retrieval}}. ACM, \bibinfo{pages}{115--124}.
\newblock


\bibitem[\protect\citeauthoryear{Liu, Ott, Goyal, Du, Joshi, Chen, Levy, Lewis,
  Zettlemoyer, and Stoyanov}{Liu et~al\mbox{.}}{2019}]%
        {liu2019roberta}
\bibfield{author}{\bibinfo{person}{Yinhan Liu}, \bibinfo{person}{Myle Ott},
  \bibinfo{person}{Naman Goyal}, \bibinfo{person}{Jingfei Du},
  \bibinfo{person}{Mandar Joshi}, \bibinfo{person}{Danqi Chen},
  \bibinfo{person}{Omer Levy}, \bibinfo{person}{Mike Lewis},
  \bibinfo{person}{Luke Zettlemoyer}, {and} \bibinfo{person}{Veselin
  Stoyanov}.} \bibinfo{year}{2019}\natexlab{}.
\newblock \showarticletitle{{RoBERTa}: A Robustly Optimized BERT Pretraining
  Approach}.
\newblock \bibinfo{journal}{\emph{arXiv preprint arXiv:1907.11692}}
  (\bibinfo{year}{2019}).
\newblock


\bibitem[\protect\citeauthoryear{Malinen and Fr{\"a}nti}{Malinen and
  Fr{\"a}nti}{2014}]%
        {malinen2014balanced}
\bibfield{author}{\bibinfo{person}{Mikko~I Malinen} {and} \bibinfo{person}{Pasi
  Fr{\"a}nti}.} \bibinfo{year}{2014}\natexlab{}.
\newblock \showarticletitle{Balanced k-means for clustering}. In
  \bibinfo{booktitle}{\emph{Joint IAPR International Workshops on Statistical
  Techniques in Pattern Recognition (SPR) and Structural and Syntactic Pattern
  Recognition (SSPR)}}. Springer, \bibinfo{pages}{32--41}.
\newblock


\bibitem[\protect\citeauthoryear{Mikolov, Sutskever, Chen, Corrado, and
  Dean}{Mikolov et~al\mbox{.}}{2013}]%
        {mikolov2013distributed}
\bibfield{author}{\bibinfo{person}{Tomas Mikolov}, \bibinfo{person}{Ilya
  Sutskever}, \bibinfo{person}{Kai Chen}, \bibinfo{person}{Greg~S Corrado},
  {and} \bibinfo{person}{Jeff Dean}.} \bibinfo{year}{2013}\natexlab{}.
\newblock \showarticletitle{Distributed representations of words and phrases
  and their compositionality}. In \bibinfo{booktitle}{\emph{Advances in neural
  information processing systems}}. \bibinfo{pages}{3111--3119}.
\newblock


\bibitem[\protect\citeauthoryear{Nam, Menc{\'\i}a, Kim, and F{\"u}rnkranz}{Nam
  et~al\mbox{.}}{2017}]%
        {nam2017maximizing}
\bibfield{author}{\bibinfo{person}{Jinseok Nam}, \bibinfo{person}{Eneldo~Loza
  Menc{\'\i}a}, \bibinfo{person}{Hyunwoo~J Kim}, {and}
  \bibinfo{person}{Johannes F{\"u}rnkranz}.} \bibinfo{year}{2017}\natexlab{}.
\newblock \showarticletitle{Maximizing Subset Accuracy with Recurrent Neural
  Networks in Multi-label Classification}. In \bibinfo{booktitle}{\emph{NIPS}}.
\newblock


\bibitem[\protect\citeauthoryear{Partalas, Kosmopoulos, Baskiotis, Artieres,
  Paliouras, Gaussier, Androutsopoulos, Amini, and Galinari}{Partalas
  et~al\mbox{.}}{2015}]%
        {partalas2015lshtc}
\bibfield{author}{\bibinfo{person}{Ioannis Partalas}, \bibinfo{person}{Aris
  Kosmopoulos}, \bibinfo{person}{Nicolas Baskiotis}, \bibinfo{person}{Thierry
  Artieres}, \bibinfo{person}{George Paliouras}, \bibinfo{person}{Eric
  Gaussier}, \bibinfo{person}{Ion Androutsopoulos},
  \bibinfo{person}{Massih-Reza Amini}, {and} \bibinfo{person}{Patrick
  Galinari}.} \bibinfo{year}{2015}\natexlab{}.
\newblock \showarticletitle{{LSHTC}: A benchmark for large-scale text
  classification}.
\newblock \bibinfo{journal}{\emph{arXiv preprint arXiv:1503.08581}}
  (\bibinfo{year}{2015}).
\newblock


\bibitem[\protect\citeauthoryear{Pennington, Socher, and Manning}{Pennington
  et~al\mbox{.}}{2014}]%
        {pennington2014glove}
\bibfield{author}{\bibinfo{person}{Jeffrey Pennington},
  \bibinfo{person}{Richard Socher}, {and} \bibinfo{person}{Christopher~D
  Manning}.} \bibinfo{year}{2014}\natexlab{}.
\newblock \showarticletitle{Glove: Global vectors for word representation}. In
  \bibinfo{booktitle}{\emph{EMNLP}}. \bibinfo{pages}{1532--1543}.
\newblock


\bibitem[\protect\citeauthoryear{Peters, Neumann, Iyyer, Gardner, Clark, Lee,
  and Zettlemoyer}{Peters et~al\mbox{.}}{2018}]%
        {peters2018deep}
\bibfield{author}{\bibinfo{person}{Matthew~E Peters}, \bibinfo{person}{Mark
  Neumann}, \bibinfo{person}{Mohit Iyyer}, \bibinfo{person}{Matt Gardner},
  \bibinfo{person}{Christopher Clark}, \bibinfo{person}{Kenton Lee}, {and}
  \bibinfo{person}{Luke Zettlemoyer}.} \bibinfo{year}{2018}\natexlab{}.
\newblock \showarticletitle{Deep contextualized word representations}. In
  \bibinfo{booktitle}{\emph{Proceedings of the 2018 Conference of the North
  {A}merican Chapter of the Association for Computational Linguistics
  (NAACL)}}.
\newblock


\bibitem[\protect\citeauthoryear{Prabhu, Kag, Harsola, Agrawal, and
  Varma}{Prabhu et~al\mbox{.}}{2018}]%
        {prabhu2018parabel}
\bibfield{author}{\bibinfo{person}{Yashoteja Prabhu}, \bibinfo{person}{Anil
  Kag}, \bibinfo{person}{Shrutendra Harsola}, \bibinfo{person}{Rahul Agrawal},
  {and} \bibinfo{person}{Manik Varma}.} \bibinfo{year}{2018}\natexlab{}.
\newblock \showarticletitle{Parabel: Partitioned label trees for extreme
  classification with application to dynamic search advertising}. In
  \bibinfo{booktitle}{\emph{WWW}}.
\newblock


\bibitem[\protect\citeauthoryear{Prabhu and Varma}{Prabhu and Varma}{2014}]%
        {prabhu2014fastxml}
\bibfield{author}{\bibinfo{person}{Yashoteja Prabhu} {and}
  \bibinfo{person}{Manik Varma}.} \bibinfo{year}{2014}\natexlab{}.
\newblock \showarticletitle{Fastxml: A fast, accurate and stable
  tree-classifier for extreme multi-label learning}. In
  \bibinfo{booktitle}{\emph{KDD}}.
\newblock


\bibitem[\protect\citeauthoryear{Radford, Narasimhan, Salimans, and
  Sutskever}{Radford et~al\mbox{.}}{2018}]%
        {radford2018improving}
\bibfield{author}{\bibinfo{person}{Alec Radford}, \bibinfo{person}{Karthik
  Narasimhan}, \bibinfo{person}{Tim Salimans}, {and} \bibinfo{person}{Ilya
  Sutskever}.} \bibinfo{year}{2018}\natexlab{}.
\newblock \showarticletitle{Improving language understanding by generative
  pre-training}.
\newblock  (\bibinfo{year}{2018}).
\newblock


\bibitem[\protect\citeauthoryear{Reddi, Kale, Yu, Holtmann-Rice, Chen, and
  Kumar}{Reddi et~al\mbox{.}}{2019}]%
        {reddi2018stochastic}
\bibfield{author}{\bibinfo{person}{Sashank~J Reddi}, \bibinfo{person}{Satyen
  Kale}, \bibinfo{person}{Felix Yu}, \bibinfo{person}{Dan Holtmann-Rice},
  \bibinfo{person}{Jiecao Chen}, {and} \bibinfo{person}{Sanjiv Kumar}.}
  \bibinfo{year}{2019}\natexlab{}.
\newblock \showarticletitle{Stochastic Negative Mining for Learning with Large
  Output Spaces}. In \bibinfo{booktitle}{\emph{AISTATS}}.
\newblock


\bibitem[\protect\citeauthoryear{Tagami}{Tagami}{2017}]%
        {tagami2017annexml}
\bibfield{author}{\bibinfo{person}{Yukihiro Tagami}.}
  \bibinfo{year}{2017}\natexlab{}.
\newblock \showarticletitle{{AnnexML}: Approximate nearest neighbor search for
  extreme multi-label classification}. In \bibinfo{booktitle}{\emph{Proceedings
  of the 23rd ACM SIGKDD international conference on knowledge discovery and
  data mining}}. \bibinfo{pages}{455--464}.
\newblock


\bibitem[\protect\citeauthoryear{Varma}{Varma}{2019}]%
        {xmc_repo}
\bibfield{author}{\bibinfo{person}{Manik Varma}.}
  \bibinfo{year}{2019}\natexlab{}.
\newblock \bibinfo{title}{The Extreme Classification Repository: Multi-label
  Datasets \& Code}.
\newblock
  \bibinfo{howpublished}{\url{http://manikvarma.org/downloads/XC/XMLRepository.html}}.
\newblock


\bibitem[\protect\citeauthoryear{Vaswani, Shazeer, Parmar, Uszkoreit, Jones,
  Gomez, Kaiser, and Polosukhin}{Vaswani et~al\mbox{.}}{2017}]%
        {vaswani2017attention}
\bibfield{author}{\bibinfo{person}{Ashish Vaswani}, \bibinfo{person}{Noam
  Shazeer}, \bibinfo{person}{Niki Parmar}, \bibinfo{person}{Jakob Uszkoreit},
  \bibinfo{person}{Llion Jones}, \bibinfo{person}{Aidan~N Gomez},
  \bibinfo{person}{{\L}ukasz Kaiser}, {and} \bibinfo{person}{Illia
  Polosukhin}.} \bibinfo{year}{2017}\natexlab{}.
\newblock \showarticletitle{Attention is all you need}. In
  \bibinfo{booktitle}{\emph{NIPS}}.
\newblock


\bibitem[\protect\citeauthoryear{Wang, Singh, Michael, Hill, Levy, and
  Bowman}{Wang et~al\mbox{.}}{2018}]%
        {wang2018glue}
\bibfield{author}{\bibinfo{person}{Alex Wang}, \bibinfo{person}{Amanpreet
  Singh}, \bibinfo{person}{Julian Michael}, \bibinfo{person}{Felix Hill},
  \bibinfo{person}{Omer Levy}, {and} \bibinfo{person}{Samuel~R Bowman}.}
  \bibinfo{year}{2018}\natexlab{}.
\newblock \showarticletitle{Glue: A multi-task benchmark and analysis platform
  for natural language understanding}.
\newblock \bibinfo{journal}{\emph{arXiv preprint arXiv:1804.07461}}
  (\bibinfo{year}{2018}).
\newblock


\bibitem[\protect\citeauthoryear{Wolf, Debut, Sanh, Chaumond, Delangue, Moi,
  Cistac, Rault, Louf, Funtowicz, and Brew}{Wolf et~al\mbox{.}}{2019}]%
        {Wolf2019HuggingFacesTS}
\bibfield{author}{\bibinfo{person}{Thomas Wolf}, \bibinfo{person}{Lysandre
  Debut}, \bibinfo{person}{Victor Sanh}, \bibinfo{person}{Julien Chaumond},
  \bibinfo{person}{Clement Delangue}, \bibinfo{person}{Anthony Moi},
  \bibinfo{person}{Pierric Cistac}, \bibinfo{person}{Tim Rault},
  \bibinfo{person}{R'emi Louf}, \bibinfo{person}{Morgan Funtowicz}, {and}
  \bibinfo{person}{Jamie Brew}.} \bibinfo{year}{2019}\natexlab{}.
\newblock \showarticletitle{HuggingFace's Transformers: State-of-the-art
  Natural Language Processing}.
\newblock \bibinfo{journal}{\emph{ArXiv}}  \bibinfo{volume}{abs/1910.03771}
  (\bibinfo{year}{2019}).
\newblock


\bibitem[\protect\citeauthoryear{Wydmuch, Jasinska, Kuznetsov, Busa-Fekete, and
  Dembczynski}{Wydmuch et~al\mbox{.}}{2018}]%
        {wydmuch2018no}
\bibfield{author}{\bibinfo{person}{Marek Wydmuch}, \bibinfo{person}{Kalina
  Jasinska}, \bibinfo{person}{Mikhail Kuznetsov}, \bibinfo{person}{R{\'o}bert
  Busa-Fekete}, {and} \bibinfo{person}{Krzysztof Dembczynski}.}
  \bibinfo{year}{2018}\natexlab{}.
\newblock \showarticletitle{A no-regret generalization of hierarchical softmax
  to extreme multi-label classification}. In \bibinfo{booktitle}{\emph{NIPS}}.
\newblock


\bibitem[\protect\citeauthoryear{Yang, Dai, Yang, Carbonell, Salakhutdinov, and
  Le}{Yang et~al\mbox{.}}{2019}]%
        {yang2019xlnet}
\bibfield{author}{\bibinfo{person}{Zhilin Yang}, \bibinfo{person}{Zihang Dai},
  \bibinfo{person}{Yiming Yang}, \bibinfo{person}{Jaime Carbonell},
  \bibinfo{person}{Ruslan Salakhutdinov}, {and} \bibinfo{person}{Quoc~V Le}.}
  \bibinfo{year}{2019}\natexlab{}.
\newblock \showarticletitle{{XLNet}: Generalized Autoregressive Pretraining for
  Language Understanding}. In \bibinfo{booktitle}{\emph{NIPS}}.
\newblock


\bibitem[\protect\citeauthoryear{Yen, Huang, Dai, Ravikumar, Dhillon, and
  Xing}{Yen et~al\mbox{.}}{2017}]%
        {yen2017ppdsparse}
\bibfield{author}{\bibinfo{person}{Ian~EH Yen}, \bibinfo{person}{Xiangru
  Huang}, \bibinfo{person}{Wei Dai}, \bibinfo{person}{Pradeep Ravikumar},
  \bibinfo{person}{Inderjit Dhillon}, {and} \bibinfo{person}{Eric Xing}.}
  \bibinfo{year}{2017}\natexlab{}.
\newblock \showarticletitle{{PPD}sparse: A parallel primal-dual sparse method
  for extreme classification}. In \bibinfo{booktitle}{\emph{KDD}}. ACM.
\newblock


\bibitem[\protect\citeauthoryear{You, Zhang, Wang, Dai, Mamitsuka, and Zhu}{You
  et~al\mbox{.}}{2019}]%
        {you2019attentionxml}
\bibfield{author}{\bibinfo{person}{Ronghui You}, \bibinfo{person}{Zihan Zhang},
  \bibinfo{person}{Ziye Wang}, \bibinfo{person}{Suyang Dai},
  \bibinfo{person}{Hiroshi Mamitsuka}, {and} \bibinfo{person}{Shanfeng Zhu}.}
  \bibinfo{year}{2019}\natexlab{}.
\newblock \showarticletitle{AttentionXML: Label Tree-based Attention-Aware Deep
  Model for High-Performance Extreme Multi-Label Text Classification}. In
  \bibinfo{booktitle}{\emph{Advances in Neural Information Processing
  Systems}}. \bibinfo{pages}{5812--5822}.
\newblock


\end{thebibliography}

%\appendix

\end{document}